\documentclass[letterpaper]{article} 
\usepackage{aaai2026}  
\usepackage{times}  
\usepackage{helvet}  
\usepackage{courier}  
\usepackage[hyphens]{url}  
\usepackage{graphicx} 
\usepackage{natbib}  
\usepackage{caption} 
\usepackage{algorithm}
\usepackage{algpseudocode}
\usepackage{amsmath, amssymb}

\newcommand{\safe}{\text{safe}}
\usepackage[font=small,captionskip=0cm]{subfig}
\usepackage{enumitem}
\usepackage{cite}
\usepackage{xcolor}
\usepackage{newfloat}
\usepackage{listings}
\DeclareCaptionStyle{ruled}{labelfont=normalfont,labelsep=colon,strut=off} 
\floatstyle{ruled}
\newfloat{listing}{tb}{lst}{}
\floatname{listing}{Listing}
\title{Formal Safety Guarantees for Autonomous Vehicles using Barrier Certificates}
\author{
   Oumaima Barhoumi\textsuperscript{\rm 1},
    Mohamed H Zaki\textsuperscript{\rm 2},
   Sofiène Tahar\textsuperscript{\rm 1}
}
\affiliations{
    \textsuperscript{\rm 1}Department of Electrical and Computer Engineering, Concordia University, Montreal, Quebec, Canada\\
    \textsuperscript{\rm 2}Department of Civil and Environmental Engineering, Western University, London, Ontario, Canada
    oumayma.barhoumi@mail.concordia.ca
}

\usepackage{bibentry}

\begin{document}

\maketitle

\begin{abstract}
	Modern AI technologies enable autonomous vehicles to perceive complex scenes, predict human behavior, and make real-time driving decisions. However, these data-driven components often operate as black boxes, lacking interpretability and rigorous safety guarantees. Autonomous vehicles operate in dynamic, mixed-traffic environments where interactions with human-driven vehicles introduce uncertainty and safety challenges. This work develops a formally verified safety framework for Connected and Autonomous Vehicles (CAVs) that integrates Barrier Certificates (BCs) with interpretable traffic conflict metrics, specifically Time-to-Collision (TTC) as a spatio-temporal safety metric. Safety conditions are verified using Satisfiability Modulo Theories (SMT) solvers, and an adaptive control mechanism ensures vehicles comply with these constraints in real time. Evaluation on real-world highway datasets shows a significant reduction in unsafe interactions, with up to 40\% fewer events where TTC falls below a 3 seconds threshold, and complete elimination of conflicts in some lanes. This approach provides both interpretable and provable safety guarantees, demonstrating a practical and scalable strategy for safe autonomous driving.
\end{abstract}

\section{Introduction}

Embodied Artificial Intelligence (AI) systems, such as autonomous vehicles, aerial drones, and mobile robots, are increasingly deployed in safety-critical contexts where errors can have physical and societal consequences. Modern AI and machine learning (ML) models have demonstrated remarkable capabilities in perception, prediction, and decision-making, enabling machines to interpret complex sensory data and adapt to dynamic environments. However, these data-driven systems remain inherently limited when it comes to guaranteeing reliable and safe operation under all conditions. First, the opacity of deep learning models prevents clear understanding of their internal decision processes, making it difficult to explain or predict their behavior in novel or rare scenarios. Second, their dependence on training data means that performance is often correlated with data coverage, when faced with unobserved conditions, such as unusual traffic patterns, weather, or sensor noise, model predictions can degrade unpredictably. Third, these systems lack rigorous safety assurances: even small perturbations or distribution shifts can lead to unsafe control actions, yet there is no mathematical framework to certify that learned policies will always remain within safe operational bounds. Finally, traditional validation methods—including extensive testing and simulation, are inherently incomplete, as they can only sample a subset of possible scenarios in high-dimensional, continuous environments.

In transportation, these limitations are particularly critical. Autonomous vehicles must operate in mixed-traffic conditions, where human-driven vehicles exhibit stochastic, non-cooperative, and sometimes adversarial behavior. Without provable safety guarantees, the integration of AI-based autonomy into real-world traffic systems introduces substantial risks. Addressing these gaps requires bridging the strengths of AI with formal reasoning techniques capable of providing interpretable, testable, and verifiable safety guarantees—an urgent challenge in the broader pursuit of trustworthy embodied AI. Formal methods are computerized mathematical reasoning tools based on logic theory capable of guaranteeing accuracy and soundness. This work contributes to that vision by combining interpretable traffic metrics, such as Time-to-Collision (TTC)~\cite{hayward1971near}, with mathematically provable constraints (Barrier Certificates, BCs)~\cite{prajna2004safety} and formal verification tools based on Satisfiability Modulo Theories (SMT)~\cite{de2008z3}. The resulting framework provides certifiable safety guarantees and supports reliable operation of AI-enabled autonomous systems, thereby complementing the perception and decision-making strengths of AI with formal verification rigor.

Recent research has explored related directions toward enforcing safety through control-theoretic certificates. In particular, several studies have integrated formal specification languages with the formulation of Control Barrier Functions (CBFs) to ensure constraint satisfaction in dynamic systems. For instance, the works in~\cite{lindemann2019decentralized,sharifi2023platoons} present control strategies for multi-agent systems where each agent is assigned a local Signal Temporal Logic (STL) task that may depend on others’ behaviors, potentially creating conflicts. Their common approach involves constructing time-varying convergent CBFs to guarantee satisfaction of STL tasks. Similarly, studies such as~\cite{lindemann2020barrier,srinivasan2020control,tian2023control} leverage barrier functions to ensure that multi-agent or robotic systems meet temporal logic specifications related to safety, reachability, or motion planning. A key limitation of these approaches lies in assuming the validity of the certificates through their construction—without formally verifying their correctness. While STL is expressive enough to capture complex temporal and task-dependent requirements, the synthesized barrier certificates may lack soundness guarantees, leaving room for modeling or implementation errors. To address this gap, we propose to leverage formal verification techniques to rigorously validate that Barrier Certificates (BC)—whether automatically synthesized or manually defined—encode true safety invariants. Specifically, we use the Z3 SMT solver~\cite{de2008z3} to formally verify the satisfiability of BC-based safety conditions. 

In this paper, we investigate car-following scenarios, where rear-end interactions are among the most critical safety scenarios, we further employ Time-to-Collision (TTC) as a spatio-temporal safety indicator. TTC naturally quantifies the risk in longitudinal vehicle interactions and thus provides an intuitive basis for constructing interpretable BC formulations, with a TTC threshold defining the safety margin. Finally, we demonstrate the practical relevance of this approach on large-scale real-world highway data, showing how data-driven AI components can be complemented by symbolic reasoning to achieve interpretable and verifiable autonomy.

\section{Proposed Methodology}
\begin{figure*}[t]
	\centering
	\includegraphics[width=0.8\textwidth]{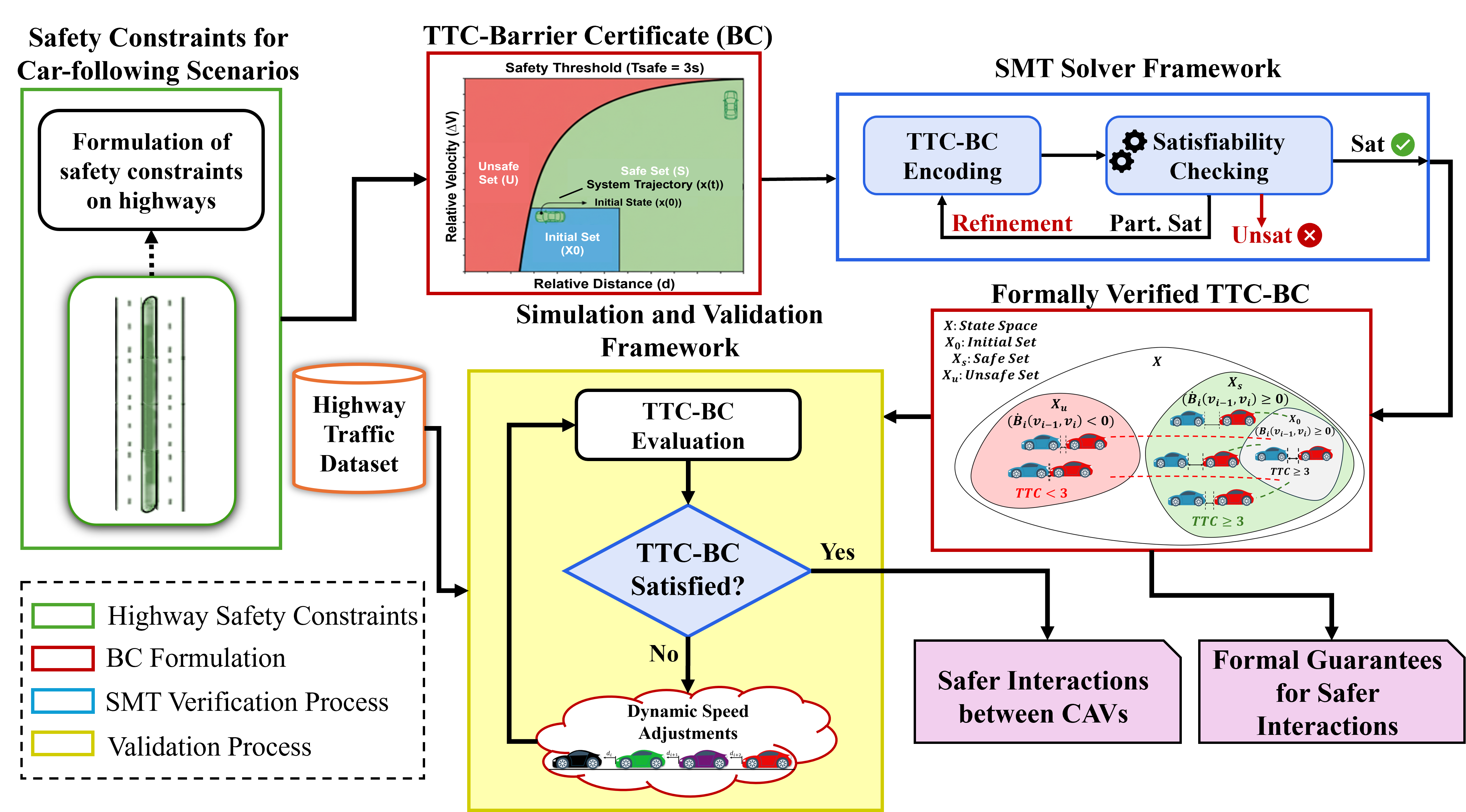}
	\caption{Formally Verified TTC-BC for Safe Interactions between CAVs on Highways}
	\label{fig:metho1}
\end{figure*}

Figure~\ref{fig:metho1} illustrates the proposed formal framework for the development, verification, and validation of safety-critical guarantees in car-following scenarios using TTC-based barrier certificates (TTC-BC).
The depicted methodology ensures safe CAV interactions on highways using formally verified TTC barrier certificates, comprising four main stages: \textit{Barrier Certificate Formulation}, \textit{Formal Verification}, \textit{Validation}, and \textit{Dynamic Speed Adjustment}. The first stage translates highway car-following safety constraints into a mathematical representation by formulating a TTC-BC, which partitions the state-space of relative distance and velocity into safe and unsafe regions, with the boundary defined by a TTC threshold. To ensure the BC holds under all conditions, formal verification is performed using the Z3 SMT solver. The BC properties are encoded as logical constraints, and the solver searches for counterexamples where a safe state could transition to an unsafe one. If no counterexamples are found, the BC is formally proven safe. In some cases, the solver may identify the constraints as \textit{partially satisfiable}, meaning that safety holds only under a subset of conditions. These cases guide the refinement of constraint definitions and boundary conditions to strengthen the safety guarantees. If, after refinement, the solver determines that no feasible solution exists, the constraint is formally \textit{unsatisfiable}, indicating that the defined safety property cannot be guaranteed under the given assumptions. Finally, the effectiveness of the verified TTC-BC is validated on a real-world highway dataset from the German Autobahn~\cite{highDdataset}. Vehicle pairs violating the BC trigger dynamic speed adjustments to return them to the safe region. This integrated approach combines formal guarantees with empirical validation to ensure safer interactions among CAVs.

\section{SMT-Based Safety Verification}
	Satisfiability Modulo Theories (SMT){~\cite{de2008z3}} is a foundational technique in formal methods that generalizes Boolean satisfiability (SAT) by supporting richer mathematical theories beyond propositional logic. Constraints are typically encoded as first-order logic formulas combined with background theories such as arithmetic, arrays or 
In our case, we encode the Time-to-Collision-based Barrier Certificate (TTC-BC), expressed as \(B(i, i-1) = \frac{x_{i-1} - x_i - L}{v_i - v_{i-1}} - T_{\text{safe}}\), to capture the safety relation between a following vehicle \(i\) and its leader \(i-1\), where \(x\) and \(v\) denote the longitudinal position and velocity of each vehicle, respectively, and \(L\) represents the vehicle length.
Using the Z3 SMT solver, we then verify whether the TTC-BC constraints remain satisfiable under nonlinear vehicle dynamics in car-following scenarios, ensuring that safety invariants hold across realistic traffic interactions. If a violation is predicted, an adaptive control process adjusts the CAV’s velocity to restore safety. To this end, we employ a counterexample-based verification approach proving that a system is safe by searching for a specific instance where it fails. This formal reasoning ensures that CAVs remain within the verified safe region under all operating conditions. Algorithm{~\ref{algo}}, given below, describes the formal safety verification process using the Z3 SMT solver.

	\begin{algorithm}[tb]
	\caption{Safety Verification using \texttt{Z3}}\label{algo}
	\small
	\begin{algorithmic}[1]
		\State \textbf{Input:} Vehicle dynamics constraints, safety parameters
		\State \textbf{Output:} Safety guarantee or counterexample
		\Statex
		\State Define symbolic variables: $N, T_{\safe}, L, x[i], v[i], a[i]$
		\State Add physical constraints: $x[i] > 0, v[i] > 0, a[i] \in [-6, 3]$
		\State Add ordering constraints: $x[i] < x[i-1], v[i] > v[i-1]$
		\State Add collision avoidance: $x[i-1] - x[i] - L > 0$
		\State \textbf{Barrier Certificate:} 
		\State $\forall i \in [1, N): B(i,i-1) \geq 0 \Rightarrow \dot{B}(i,i-1) \geq 0$
		\State \textbf{Verification Query:}
		\State $\exists i \in [1, N): B(i,i-1) \geq 0 \land \dot{B}(i,i-1) < 0$
		\If{query is satisfiable}
		\State \Return Counterexample found (System unsafe)
		\Else
		\State \Return Safety verified (System safe)
		\EndIf
	\end{algorithmic}
\end{algorithm}

\section{Case Study: Validation using the HighD Dataset}

\begin{figure}[t]
	\centering
	\includegraphics[width=0.9\columnwidth]{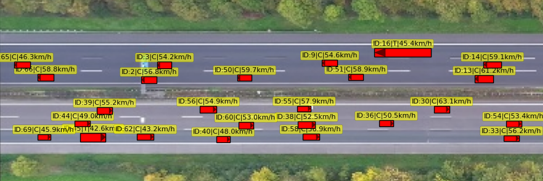} 
	\caption{Visualization of car-following behavior on a Highway Segment.}
	\label{fig:track25}
\end{figure}

The framework was validated using a real-world highway dataset, called HighD, representing a large-scale collection of naturalistic vehicle trajectories from German highways as shown by Figure~\ref{fig:track25}. This data allows quantitative evaluation of safety improvement. By applying our formally verified TTC-based barrier certificates to the diverse car-following scenarios within this dataset, we can rigorously test whether our approach successfully prevents collisions under real-world conditions. The use of this dataset allows us to move beyond theoretical proofs and simulations to demonstrate the effectiveness and robustness of our formal barrier certificate in a realistic and challenging environment.

\begin{figure}[t]
	\centering
	\subfloat[
	\label{fig:nbr_conf_300}]{
		\includegraphics[width=0.9\columnwidth]{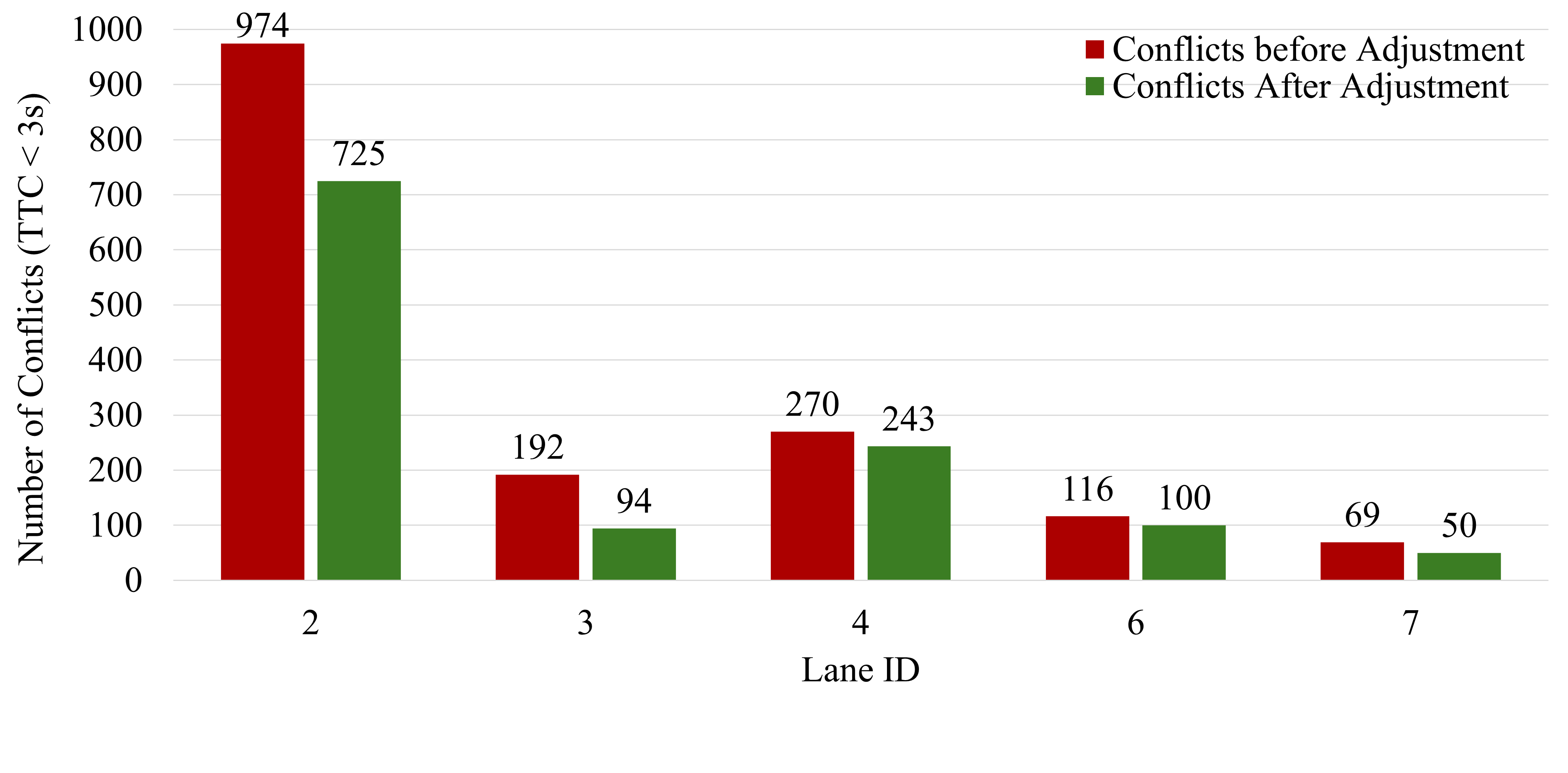}
	}
	\hfill
	\subfloat[
	\label{fig:nbr_conf_3000}]{
		\includegraphics[width=0.9\columnwidth]{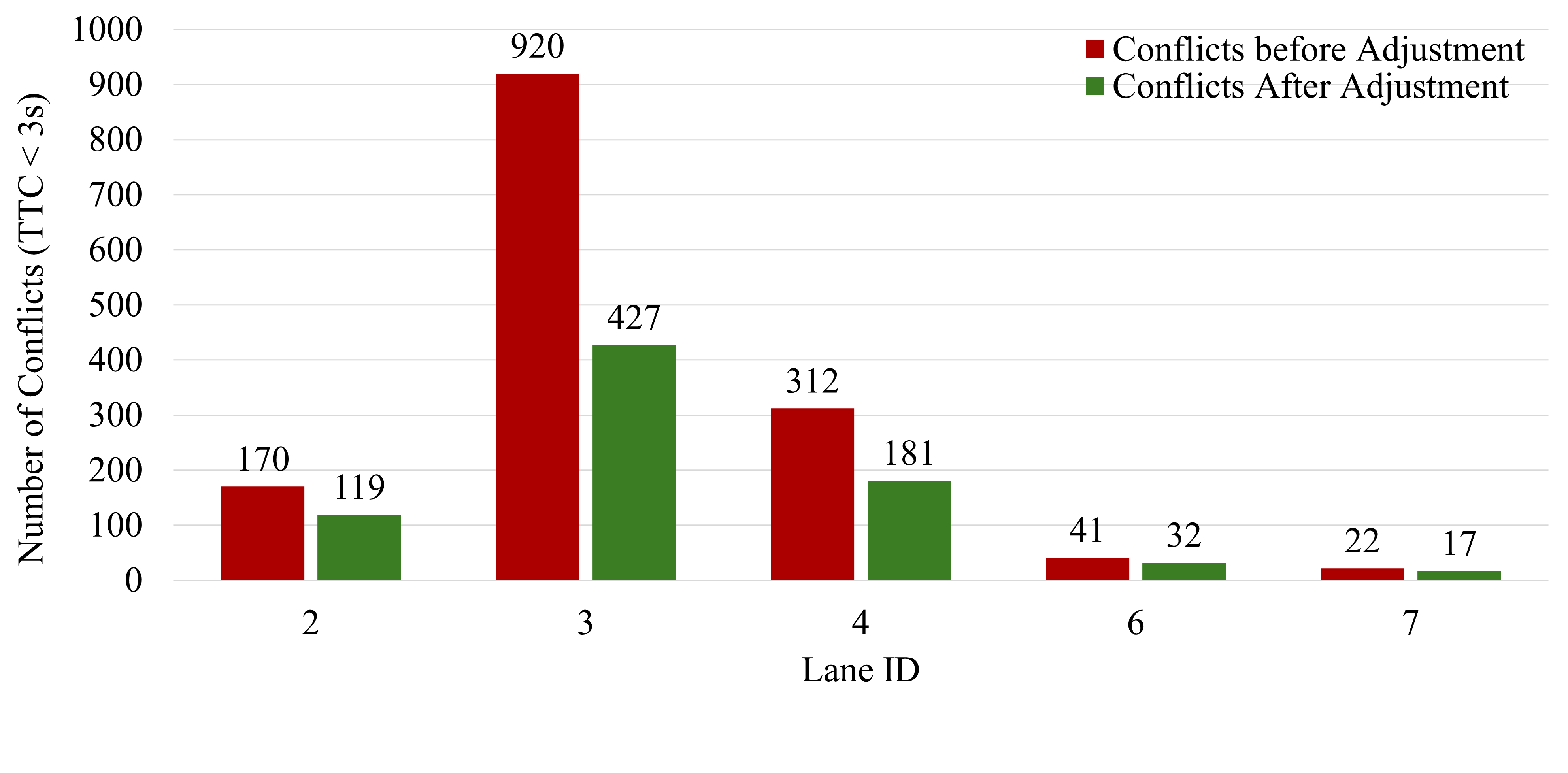}
	}
	\caption{Traffic Conflicts Comparison: Before vs. After Speed Adjustment}
	\label{fig:nbr_conf}
\end{figure}

	To resolve a traffic conflict and restore safety, a driver or an autonomous system must make speed adjustments. In a car-following scenario, the primary adjustment is deceleration. By braking, the following vehicle increases its distance from the lead vehicle, which in turn increases the TTC and provides a safer temporal buffer. In this context, we propose a speed adjustment strategy for vehicles identified within the unsafe region (i.e., those with $TTC < T_{\text{safe}}$). The adjustment consists of reducing the velocity of the violating vehicles so that their relative motion with respect to the leading vehicle results in a longer available reaction time. After applying this adjustment, the TTC values are recalculated, and the results consistently show an increase, confirming a recovery of safety margins. Across multiple traffic segments, the number of potential conflicts—defined as instances with $TTC < 3\,\text{s}$, was reduced by up to 40\% after applying BC-based regulation as depicted by Figure~\ref{fig:nbr_conf} for 300 (a) and 3000 (b) frames, respectively. In certain lanes, unsafe interactions were entirely eliminated. The approach thus provides both quantitative safety gains and qualitative interpretability: drivers and engineers can directly interpret TTC thresholds, while formal verification ensures mathematically provable safety invariance. Combining data-driven testing with formal verification provides a practical and scalable approach for safe autonomous systems, connecting formal guarantees with real-world operation.

	\section{Conclusion}
This research presents a methodology for ensuring the safety of autonomous vehicles in car-following scenarios by integrating Time-To-Collision (TTC)-based Barrier Certificates (BC), formal verification, and real-world data validation. The BC rigorously defines safe operating regions and is formally verified using an SMT solver, providing provable guarantees beyond traditional simulation. Validation on the HighD highway dataset demonstrates that dynamic speed adjustments guided by TTC effectively mitigate traffic conflicts. Overall, this approach offers a rigorous and practical framework for certifying safety-critical autonomous vehicle controllers in complex traffic environments.

\bibliography{aaai2026}

\end{document}